\documentclass[journal]{IEEEtran}

\ifCLASSINFOpdf
\else
\fi

\usepackage{times}
\usepackage{epsfig}
\usepackage{graphicx}
\usepackage{amsmath}
\usepackage{amssymb}
\usepackage{subfig}
\usepackage{multirow}
\usepackage{booktabs}
\usepackage{hyperref}

\usepackage{float} 

\newlength{\w}

\graphicspath{{figures/}}
\DeclareGraphicsExtensions{.eps,.png,.pdf,.jpg,.JPG}

\begin{document}
%
\title{DeepFakes: a New Threat to Face Recognition? Assessment and Detection}

%
%
%

\author{Pavel Korshunov and
        S\'{e}bastien Marcel
\thanks{P. Korshunov and S. Marcel are at Idiap Research Institute,
Martigny, Switzerland; emails: \{pavel.korshunov,sebastien.marcel\}@idiap.ch}
}

\maketitle

\begin{abstract}

It is becoming increasingly easy to automatically replace a face of one person in a video with the face of another person by using a pre-trained generative adversarial network (GAN). 
Recent public scandals, e.g., the faces of celebrities being swapped onto pornographic videos, call for automated ways to detect these Deepfake videos. 
To help developing such methods, in this paper, we present the first publicly available set of Deepfake videos generated from videos of VidTIMIT database. We used open source software based on GANs to create the Deepfakes, and we emphasize that training and blending parameters can significantly impact the quality of the resulted videos. 
To demonstrate this impact, we generated videos with low and high visual quality (320 videos each) using differently tuned parameter sets. We showed that the state of the art face recognition systems based on VGG and Facenet neural networks are vulnerable to Deepfake videos, with 85.62\% and 95.00\% false acceptance rates (on high quality versions) respectively, which means methods for detecting Deepfake videos are necessary.
By considering several baseline approaches, we found that audio-visual approach based on lip-sync inconsistency detection was not able to distinguish Deepfake videos. The best performing method, which is based on visual quality metrics and is often used in presentation attack detection domain, resulted in 8.97\% equal error rate on high quality Deepfakes. Our experiments demonstrate that GAN-generated Deepfake videos are challenging for both face recognition systems and existing detection methods, and the further development of face swapping technology will make it even more so.


\end{abstract}

\begin{IEEEkeywords}
Deepfake videos, Face swapping, Video database, Tampering detection, Face recognition
\end{IEEEkeywords}

%
\IEEEpeerreviewmaketitle

\section{Introduction}
\label{sec:intro}

Recent advances in automated video and audio editing tools, generative adversarial networks (GANs), and social media allow creation and fast dissemination of high quality tampered video content. Such content already led to appearance of deliberate misinformation, coined `fake news', which is impacting political landscapes of several countries~\cite{allcott_gentzkow_2017}. A recent surge of videos, often obscene, in which a face can be swapped with someone else's using neural networks, so called Deepfakes\footnote{{\scriptsize Open source:} \url{https://github.com/deepfakes/faceswap}}, are of a great public concern\footnote{{\scriptsize BBC report (Feb 3, 2018):} \url{http://www.bbc.com/news/technology-42912529}}. 
Accessible open source software and apps for such face swapping lead to large amounts of synthetically generated Deepfake videos appearing in social media and news, posing a significant technical challenge for detection and filtering of such content. 
Therefore, the development of efficient tools that can automatically detect these videos with swapped faces is of a paramount importance. 

Until recently, most of the research was focusing on advancing the face swapping technology~\cite{Isola2016,Korshunova2017,Nirkin2018,Pham2018}. However, responding to the public demand to detect face swapping technology, researchers are starting to work on databases and detection methods, including image and video data~\cite{Verdoliva2018} generated with an older face swapping approach Face2Face~\cite{Thies2016} or videos collected using Snapchat\footnote{\url{https://www.snapchat.com/}} application~\cite{Agarwal2017}. 

\begin{figure*}[tbh]
\centering
\setlength{\w}{0.16\textwidth}
\subfloat{\includegraphics[width=\w]{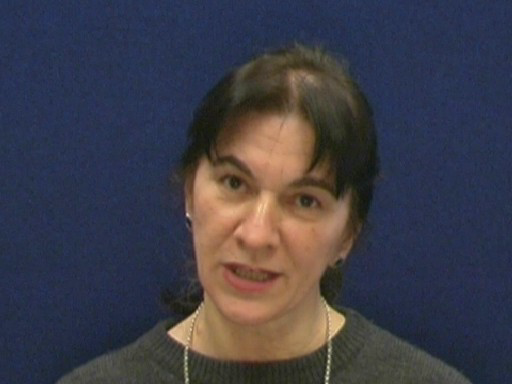}}
\subfloat{\includegraphics[width=\w]{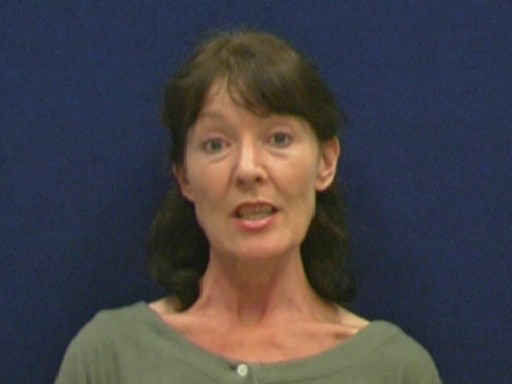}}
\subfloat{\includegraphics[width=\w]{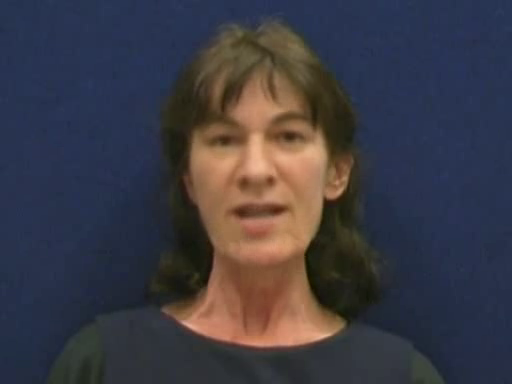}}
\subfloat{\includegraphics[width=\w]{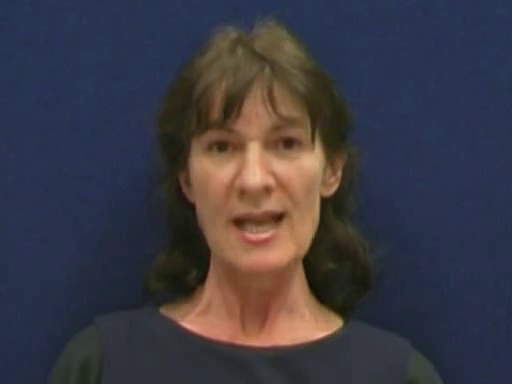}}
\subfloat{\includegraphics[width=\w]{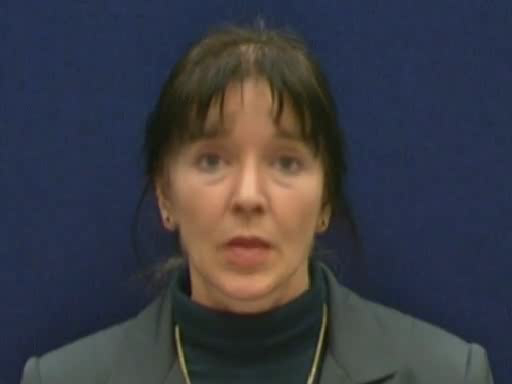}}
\subfloat{\includegraphics[width=\w]{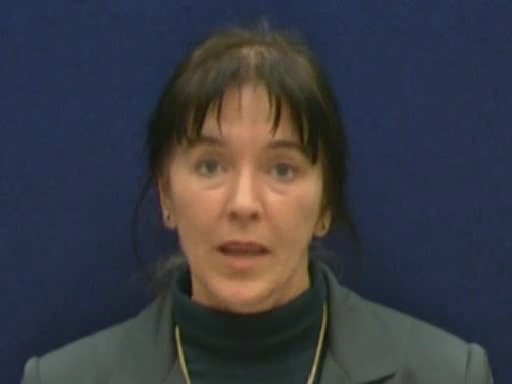}} \\
\subfloat[Original 1]{\includegraphics[width=\w]{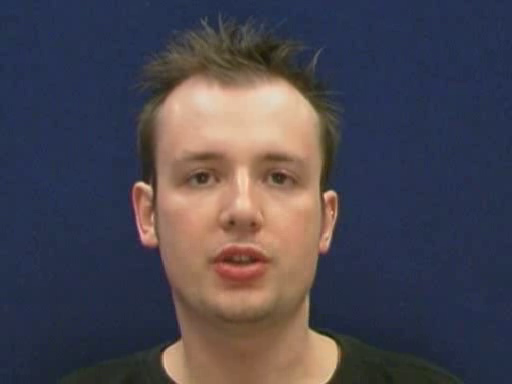}\label{fig:source}}
\subfloat[Original 2]{\includegraphics[width=\w]{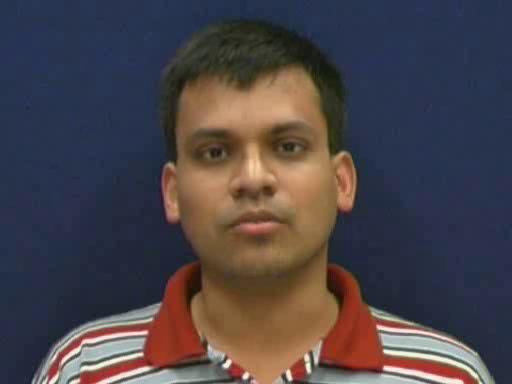}\label{fig:target}}
\subfloat[LQ swap $1\rightarrow 2$]{\includegraphics[width=\w]{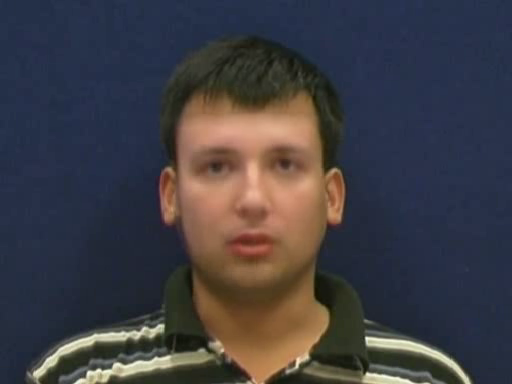}\label{fig:swaplq1}}
\subfloat[HQ swap $1\rightarrow 2$]{\includegraphics[width=\w]{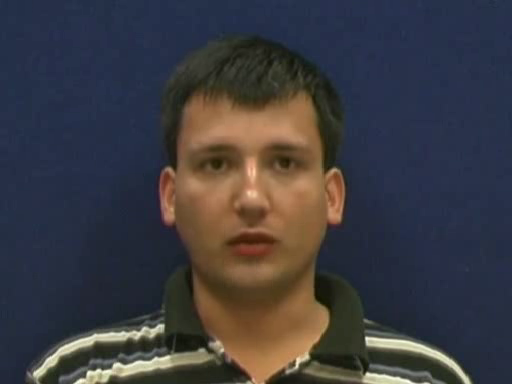}\label{fig:swaphq1}}
\subfloat[LQ swap $2\rightarrow 1$]{\includegraphics[width=\w]{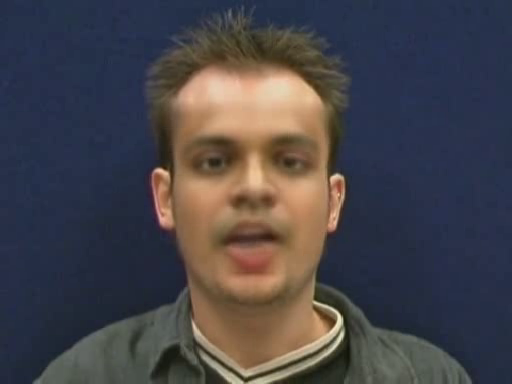}\label{fig:swaplq2}}
\subfloat[HQ swap $2\rightarrow 1$]{\includegraphics[width=\w]{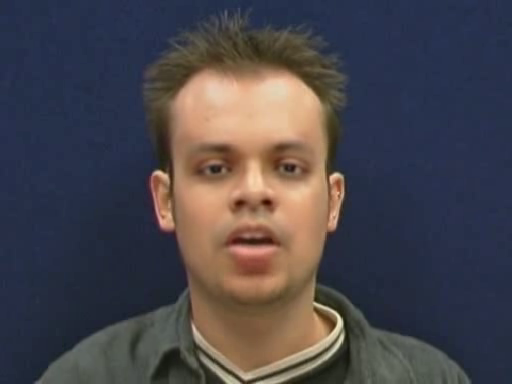}\label{fig:swaphq2}} \\
  
\caption{Screenshot of the original videos from VidTIMIT database and low (LQ) and high quality (HQ) Deepfake videos.}
\label{fig:examples}
\end{figure*}

In this paper, we present a first publicly available database of videos where faces are swapped using the open source GAN-based approach\footnote{\url{https://github.com/shaoanlu/faceswap-GAN}}, 
which is developed from the original autoencoder-based Deepfake algorithm\footnotemark[1]. We manually selected $16$ similar looking pairs of people from publicly available VidTIMIT database\footnote{\url{http://conradsanderson.id.au/vidtimit/}}. For each $32$ subject, we trained two different models, in the paper, referred to as low quality (LQ), with $64 \times 64$ input/output size, and high quality (HQ), with $128 \times 128$ size, models (see Figure~\ref{fig:examples} for examples). Since there are $10$ videos per person in VidTIMIT database, we generated $320$ videos corresponding to each version, resulting in total $620$ videos with faces swapped. For the audio, we kept the original audio track of each video, i.e., no manipulation was done to the audio channel. 

It is also important to understand how much of a threat Deepfake videos are to face recognition systems. Because if these systems are not fooled by Deepfakes, creating a separate system for detecting Deepfakes would not be necessary. To the vulnerability of face recognition to Deepfake videos, we evaluate two state of the art systems: based on VGG~\cite{Parkhi15} and Facenet\footnote{\url{https://github.com/davidsandberg/facenet}}~\cite{Schroff2015} neural networks on both untampered videos and videos with faces swapped. 

For detection of the Deepfakes, we first used an audio-visual approach that detects inconsistency between visual lip movements and speech in audio~\cite{Korshunov2018a}. It allows us to understand how well the generated Deepfakes can mimic mouth movement and whether the lips are synchronized with the speech. We also applied several baseline methods from presentation attack detection domain, by treating Deepfake videos as digital presentation attacks~\cite{Agarwal2017}, including simple principal component analysis (PCA) and linear discriminant analysis (LDA) approaches, and the approach based on image quality metrics (IQM) and support vector machine (SVM)~\cite{Galbally2014,Wen2015}.

To allow researchers to verify, reproduce, and extend our work, we provide the database of Deepfake videos\footnote{\url{https://www.idiap.ch/dataset/deepfaketimit}}, face recognition and Deepfake detection systems with corresponding scores as an open source Python package\footnote{{\scriptsize Complete implementation:} \url{https://gitlab.idiap.ch/bob/bob.report.deepfakes}}.

Therefore, this paper has the following main contributions:
\begin{itemize}
\item {Publicly available database of low and high quality sets of videos from VidTIMIT database with swapped faces using GAN-based approach;}
\item {Vulnerability analysis of VGG and Facenet based face recognition systems;}
\item {Evaluation of several detection methods of Deepfakes, including lip-syncing approach and image quality metrics with SVM method;}
\end{itemize}

\section{Related work}
\label{sec:related}

%

One of the first works on face swapping is by Bitouk~\emph{et al.}~\cite{Bitouk2008}, where the authors searched in a database for a face similar in appearance to the input face and then focused on perfecting the blending of the found face into the input image. The main motivation for this work was de-identification of an input face and its privacy preservation. Hence, the approach did not allow for a seamless swapping of any two given faces. Until the latest era of neural networks, most of the techniques for face swapping or facial reenacment were based on similarity searchers between faces or face patches in target and source video and various blending techniques~\cite{Arar2011,Xingjie2014,Uyl2015,Mahajan2017,Nirkin2018}.

The first approach that used a generative adversarial network to train a model between pre-selected two faces was proposed by Korshunova~\emph{et al.} in 2017~\cite{Korshunova2017}. Another related work with even a more ambitious idea was to use long short term memory (LSTM) based architecture to synthesize a mouth feature solely from an audio speech~\cite{Shlizerman2017}. Right after these publication became public, they attracted a lot of publicity. Open source approaches replicating these techniques started to appear, which resulted in the Deepfake phenomena. 

The rapid spread of Deepfakes and the ease of generating such videos are calling for a reliable detection method. So far, however, there are only few publications focusing on detecting GAN-generated videos with swapped faces and very little data for evaluation and benchmarking is publicly available. For instance, Zhang~\emph{et al.}~\cite{Zhang2017} proposed the method based on speeded up robust features (SURF) descriptors and SVM classifier. The authors evaluated this approach on a set of images where the face of one person was replaced with a face of another by applying color correction and smoothing techniques based on Gasussian blurring, which means the facial expressions of the input faces were not preserved. Another method based on LBP-like features with SVM classifier was proposed by Agarwal~\emph{et al.}~\cite{Agarwal2017} and evaluated on the videos collected by the authors with Snapchat\footnotemark[3] phone application. Snapchat uses active 3D model to swap faces in real time, so the resulted videos are not really Deepfakes, but it is still a widely used tool and database of such videos, if it will ever become public (the authors promised to release it but have not done so at the moment of publication), it can be interesting to research community.

R\"{o}ssler~\emph{et al.}~\cite{Verdoliva2018}  presented the most comprehensive database of non-Deepfake swapped faces ($500'000$ images from more than $1000$ videos)  to date. The authors also benchmarked the state of the art forgery classification and segmentation methods. The authors used Face2Face~\cite{Thies2016} tool to generate the database, which is based on expression transformation using 3D facial model and a pre-computed database of mouth interiors. 
One of the latest approaches~\cite{Li2018} proposed to use blinking detection as the means to distinguish swapped faces in Deepfake videos. The authors generated $49$ videos (not publicly available) and argued that the proposed eye blinking detection was effective in detecting Deepfake videos.

However, no public Deepfake video database where GAN-based approach was applied is available. Hence, it is unclear whether the above methods would be effective in detecting such faces. In fact, the Deepfakes that we have generated can effectively mimic the facial expressions, mouth movements, and blinking, so the current detection approaches need to be evaluated on such videos. For instance, it is practically impossible to evaluate the methods proposed in~\cite{Verdoliva2018} and~\cite{Li2018} as their implementations are not yet available.

\section{Deepfake database}
\label{sec:dataset}

As original data, we took video from VidTIMIT database\footnotemark[5].
The database contains $10$ videos for each of $43$ subjects, which were shot in controlled environment with people facing camera and reciting predetermined short phrases. From these $43$ subject, we manually selected $16$ pairs in such a way that subjects in the same pair have similar prominent visual features, e.g., mustaches or hairs styles. Using GAN-based face-swapping algorithm based on the available code\footnotemark[4], for each pair, we generated videos with swapped faces from subject one to subject two and visa versa (see Figure~\ref{fig:examples} for the video screenshots). 


For each pair of subjects, we have trained two different GAN models and generated two versions of the videos:

\begin{enumerate}
\item {The low quality (LQ) model has input and output image (facial regions only) of size $64 \times 64$. About $200$ frames from the videos of each subject were used for training and the frames were extracted at $4$ fps from the original videos. The training was done for $10'000$ iterations and took about $4$ hours per model on Tesla P40 GPU.}

\item{The high quality (HQ) model has input/output image size of $128 \times 128$. About $400$ frames extracted at $8$ fps from videos were used for training, which was done for $20'000$ iterations (about $12$ hours on Tesla P40 GPU).}
\end{enumerate}

Also, different blending techniques were used for different models. For LQ model, for each frame, a face was generated using a frame from a target video as an input. Then a facial mask was detected using a CNN-based face segmentation algorithm proposed in~\cite{Nirkin2018}. Using this mask, the generated face was blended with the face in the target video. For HQ model, the blending was done based on facial landmarks alignment between generated face and the original face in the target video. Landmarks detection was done using publicly available pre-trained MTCNN model~\cite{Zhang2016}. Finally, histogram normalization was applied when blending generated face into the target video to adjust for the lighting conditions.

\begin{figure*}[tbh]
\centering
\setlength{\w}{0.27\textwidth}
\subfloat[VGG-based face recognition]{\includegraphics[width=\w]{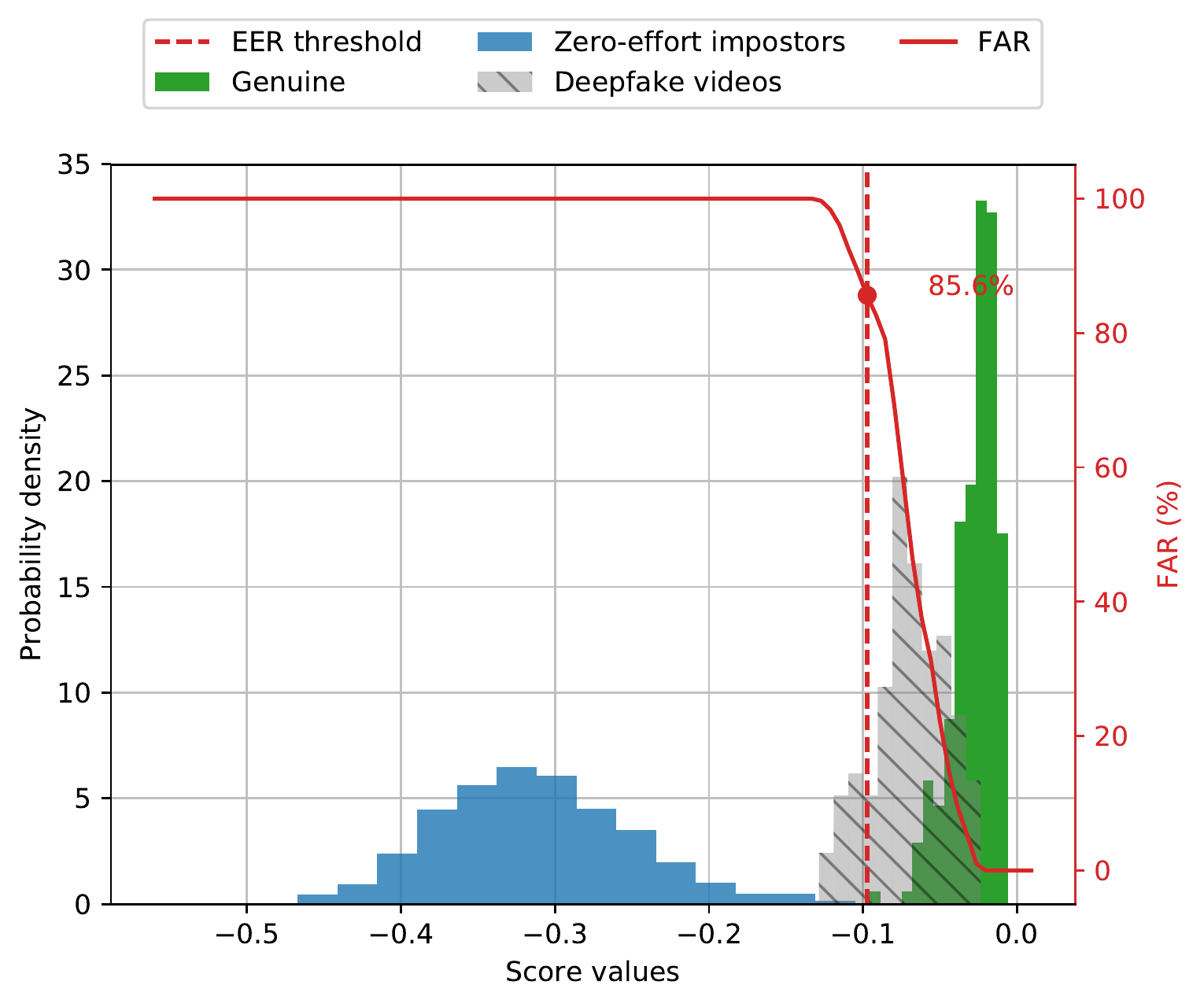}
  \label{fig:vulnvgg}}
\subfloat[FaceNet-based face recognition]{\includegraphics[width=\w]{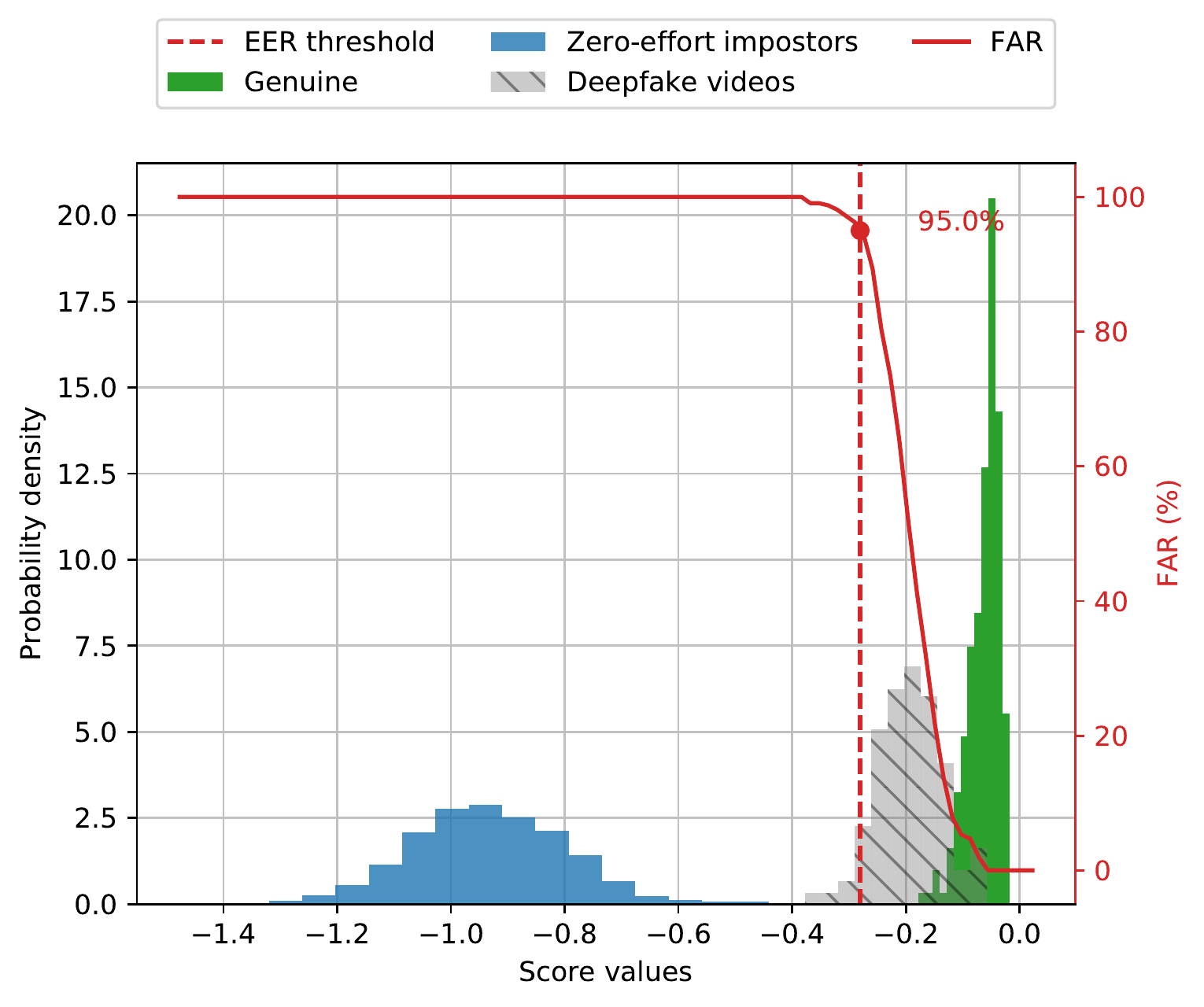}
  \label{fig:vulnfacenet}}
\subfloat[IQM+SVM Deepfake detection]{\includegraphics[width=\w]{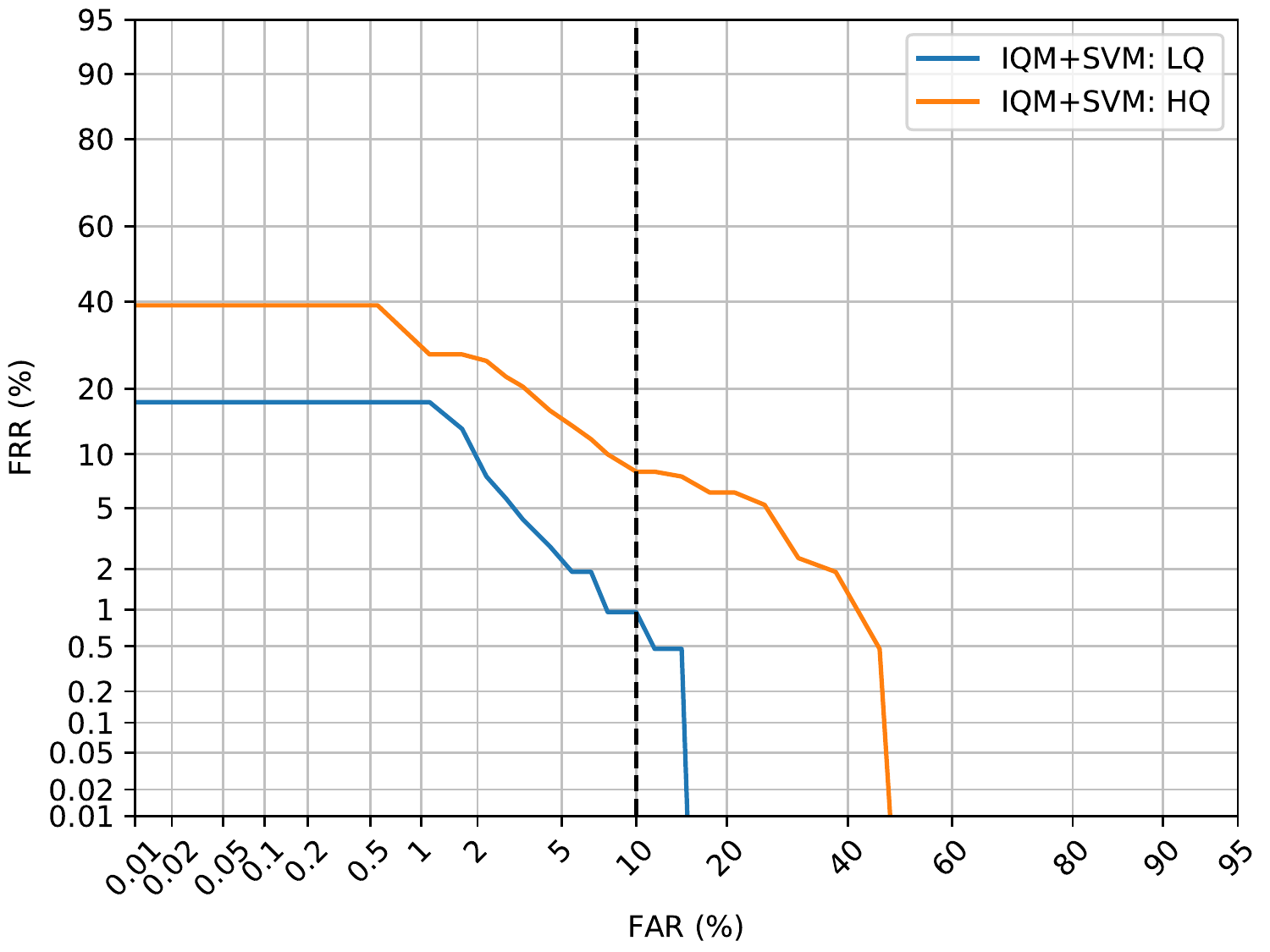}
  \label{fig:det}}
\caption{Histograms showing the vulnerability of VGG and Facenet based face recognition to high quality face-swapping and the performance of IQM+SVM detection on low and high quality Deepfakes.}
\label{fig:vuln}
\end{figure*}


\subsection{Evaluation protocol}
\label{sec:protocol}

When evaluating vulnerability of face recognition, for the \textit{licit} non-tampered scenario, we used the original VidTIMIT videos for the $32$ subjects for which we have generated corresponding Deepfake videos. In this scenario, we used $2$ videos of the subject for enrollment and the other $8$ videos as probes, for which we computed the verification scores. 

From the scores, for each possible threshold $\theta$, we computed commonly used metrics for evaluation of classification systems: false acceptance rate (FAR) and false reject rate (FRR).
Threshold at which these FAR and FRR are equal leads to an equal error rate (EER), which is commonly used as a single value metric of the system performance.

To evaluate vulnerability of face recognition to Deepfake videos, in \textit{tampered} scenario, we use these videos ($10$ for each of $32$ subjects) as probes and compute the corresponding scores using the same enrollment model as in \textit{licit} scenario. To understand if face recognition perceives Deepfakes as similar to the genuine original videos, we report the FAR metric computed using EER threshold $\theta$ from \textit{licit} scenario. If FAR value for Deepfake tampered videos is significantly higher than the one computed in \textit{licit} scenario, it means the face recognition system cannot distinguish tampered videos from originals and is therefore vulnerable to Deepfakes.

When evaluating Deepfake detection, we consider it as a binary classification problem and evaluate the ability of detection approaches to distinguish original videos from Deepfake videos. All videos in the dataset, including genuine and tampered parts, were split into training (\textit{Train}) and evaluation (\textit{Test}) subsets. To avoid bias during training and testing, we arranged that the same subject would not appear in both sets. We did not introduce a development set, which is typically used to tune hyper parameters such as threshold, because the dataset is not large enough. Therefore, we report the EER and the FRR (using the threshold when $FAR=10\%$) values on the \textit{Test} set.

\section{Analysis of deepfake videos}

In this section, we evaluate the vulnerability of face VGG~\cite{Parkhi15} and Facenet\footnote{\url{https://github.com/davidsandberg/facenet}}~\cite{Schroff2015} based recognition systems to videos with swapped faces and apply several baseline systems for detection of such videos.

\subsection{Vulnerability of face recognition}
\label{sec:vuln}

We used publicly available pre-trained VGG and Facenet architectures for face recognition. We used the \textit{fc7} and \textit{bottleneck} layers of these networks, respectively, as features and used cosine distance as a classifier. For a given test face, the confidence score of whether it belongs to a pre-enrolled model of a person is the cosine distance between the average feature vector, i.e., model, and the features vector of a test face. Both of these systems are state of the art recognition systems with VGG of $98.95\%$~\cite{Parkhi15} and Facenet of $99.63\%$~\cite{Schroff2015} accuracies on labeled faces in the wild (LFW) dataset.


We conducted the vulnerability analysis of VGG and Facenet-based face recognition systems on low quality (LQ) and high quality (HQ) face swaps in VidTIMIT database. In a \textit{licit} scenario when only original non-tampered videos are present, both systems performed very well, with EER value of $0.03\%$ for VGG and $0.00\%$ for Facenet-based system. Using the EER threshold from \textit{licit} scenario, we computed FAR value for the scenario when Deepfake videos are used as probes. In this case, for VGG the FAR is $88.75\%$ LQ Deepfakes and $85.62\%$ for HQ Deepfakes, and for Facenet the FAR is $94.38\%$ and $95.00\%$ for LQ and HQ Deepfakes respectively. To illustrate this vulnerability, we plot the score histograms for high quality Deepfake videos in Figure~\ref{fig:vuln}. 

From the results, it is clear that both VGG and Facenet based systems cannot effectively distinguish GAN-generated and swapped faces from the original ones. The fact that more advanced Facenet system is more vulnerable is also consistent with the previous findings~\cite{Mohammadi2018}. 



%
%
%
%
%

\subsection{Detection of Deepfake videos}
\label{sec:detection}

We considered several baseline Deepfake detection systems, including system that uses audio-visual data to detect inconsistencies between lip movements and audio speech, as well as, several variations of solely image based systems. 

The goal of the lip-sync based detection system is to distinguish genuine video, where lip movement and speech are synchronized, from tampered video, where lip movements and audio, which may not necessarily be speech, are not synchronized. The stages of such system include feature extraction from video and audio modalities, processing these features, and then, a two-class classifier trained to separate tampered videos from genuine. In this system, we used MFCCs as audio features~\cite{Odobez2016} and distances between mouth landmarks as visual features (inspired by~\cite{Shlizerman2017}). PCA is applied to the joint audio-visual features to reduce the dimensionality of the blocks of features and long short-term memory (LSTM)~\cite{Zisserman2016} network is trained to separate tampered and non-tampered videos as proposed in~\cite{Korshunov2018a}.

As image based systems, we implemented the following:

\begin{itemize}
\item \textit{Pixels+PCA+LDA}: use raw faces as features with PCA-LDA classifier, with $99\%$ retained variance resulting in $446$ dimensions of transform matrix.
\item \textit{IQM+PCA+LDA}: IQM features with PCA-LDA classifier with $95\%$ retained variance resulting in $2$ dimensions of transform matrix.
\item \textit{IQM+SVM}: IQM features with SVM classifier, each video has an averaged score from $20$ frames.
\end{itemize}

The systems based on image quality measures (IQM) are borrowed from the domain of presentation (including replay attacks) attack detection, where such systems have shown good performance~\cite{Galbally2014,Wen2015}. As IQM feature vector, we used $129$ measures of image quality, which include such measures like signal to noise ratio, specularity, bluriness, etc.,  by combining the features from~\cite{Galbally2014} and~\cite{Wen2015}. 

The results for all detection systems are presented in Table~\ref{tab:detection}. Figure~\ref{fig:det} shows the detection error tradeoff (DET) curves for the best performing IQM+SVM system applied to two different face swapping versions. The results demonstrate that first, lip-syncing based algorithm is not able to detect face swapping, as GANs are able to generate facial expressions with high quality that can match audio speech. Therefore, currently, only image based approaches are capable to effectively detect Deepfake videos. Second, the IQM+SVM system has a reasonably high accuracy of detecting Deepfake videos, although videos generated with HQ model pose a more serious challenge. It means that a more advanced techniques for face swapping will be even more challenging to detect.


\begin{table}[tb]
\footnotesize
\caption{Baseline detection systems for low (LQ) and high quality (HQ) Deepfake videos of VidTIMIT database. EER and FRR when FAR equal to 10\% are computed on Test set.}
\label{tab:detection}
\centering

\setlength\tabcolsep{1.5pt}
\def\arraystretch{1.3}%

\begin{tabular}{l|c|c|c}
\toprule

{\bf Database} & {\bf Detection system} & {\bf EER (\%)} & {\bf FRR@FAR10\% (\%)} \\ \midrule

& LSTM lip-sync~\cite{Korshunov2018a} & 41.8 & 81.67\\
LQ Deepfake & Pixels+PCA+LDA & 39.48 & 78.10 \\
& IQM+PCA+LDA & 20.52 & 66.67\\
& IQM+SVM & 3.33 & 0.95 \\
\midrule
HQ Deepfake & IQM+SVM & 8.97 & 9.05 \\
\bottomrule
\end{tabular}
\end{table}

%

\section{Conclusion}
In this paper, we presented a first publicly available database of $620$ Deepfake videos for $16$ pairs of subjects from VidTIMIT database. 
We generated two versions of the videos for each subject: based on low quality $64 \times 64$ GAN model and higher quality $128 \times 128$ model. We also demonstrated that state of the art VGG and Facenet-based face recognition algorithms are vulnerable to the Deepfake videos and fail to distinguish such videos from the original ones with up to $95.00\%$ equal error rate. We also evaluated several baseline face swap detection algorithms and found that lip-sync based approach fails to detect mismatches between lip movement and speech. The techniques based on image quality measures with SVM classifier can detect HQ Deepfake videos with $8.97$\% equal error rate. 

However, the continued advancements in development of face swapping techniques will result in more challenging Deepfake videos, which will be harder to detect by the existing algorithms. Therefore, new databases and new more generic methods need to be developed in the future. Subjective evaluations to study the vulnerability of human subjects to Deepfakes are also needed. Possibly, a new arms race between Deepfake methods and detection algorithms has begun.


\section*{Acknowledgements}
This research was sponsored by Hasler Foundation's VERIFAKE project and the United States Air Force, Air Force
Research Laboratory (AFRL) and the Defense Advanced Research Projects Agency
(DARPA) under Contract No.\ FA8750-16-C-0170.  The views, opinions and/or
findings expressed are those of the author and should not be interpreted as representing the official views or policies of AFRL or DARPA.

\ifCLASSOPTIONcaptionsoff
  \newpage
\fi



\bibliographystyle{IEEEtran}
\bibliography{references_pavel}
%
%
%
%

\end{document}